\documentclass[conference,compsoc]{IEEEtran}

\ifCLASSOPTIONcompsoc
  
\else
  \usepackage{cite}
\fi

\usepackage{amsmath}
\usepackage{amsthm}
\usepackage{amsfonts}
\usepackage{amssymb}

\usepackage{graphicx}
\usepackage{xcolor}
\usepackage{graphics}
\usepackage{subfigure}
\usepackage[noend]{algpseudocode}
\usepackage{algorithm}
\newcommand\NoDo{\renewcommand{\algorithmicdo}{}}
 \newcommand\NoThen{\renewcommand{\algorithmicthen}{}}

\DeclareMathOperator*{\argmax}{\arg\!\max}

\begin{document}
%
\title{Self-supervised Robust Object Detectors from Partially Labelled Datasets}

\author{\IEEEauthorblockN{Mahdieh Abbasi and Denis Laurendeau}
\IEEEauthorblockA{Department of Electrical and Computer Engineering\\ Universit\'e Laval, Qu\'ebec, Canada\\
mahdieh.abbasi.1@ulaval.ca \\ Denis.Laurendeau@gel.ulaval.ca }
\and
\IEEEauthorblockN{Christian Gagn\'e}
\IEEEauthorblockA{Mila, Canada CIFAR AI Chair, Qu\'ebec, Canada\\
Christian.Gagne@gel.ulaval.ca }
}


%


\maketitle

\begin{abstract}

In the object detection task, merging various datasets from similar contexts but with different sets of Objects of Interest (OoI) is an inexpensive way (in terms of labor cost) for crafting a large-scale dataset covering a wide range of objects. Moreover, merging datasets allows us to train one integrated object detector, instead of training several ones, which in turn resulting in the reduction of computational and time costs. However, merging the datasets from similar contexts causes the samples with partial labeling as each constituent dataset is originally annotated for its own set of OoI and ignores to annotate those objects that are become interested after merging the datasets. With the goal of training \emph{one integrated robust object detector with high generalization performance}, we propose a training framework to overcome the missing-label challenge of the merged datasets. More specifically, we propose a computationally efficient self-supervised framework to create on-the-fly pseudo-labels for the Unlabeled Positive Instances (UPIs) in the merged dataset in order to train the object detector jointly on both ground truths and pseudo labels. We evaluate our proposed framework for training YOLO on a simulated merged dataset with missing rate $\approx\!48\%$ using VOC2012 and VOC2007. We empirically show that generalization performance of YOLO trained on both ground truths and the pseudo-labels that are created by our method is $4\%$ (on average) higher than the ones trained only with the ground truth labels of the merged dataset.
\end{abstract}


%
\IEEEpeerreviewmaketitle

\section{Introduction}

Modern CNN-based object detectors such as faster R-CNN~\cite{ren2015faster} and YOLO~\cite{redmon2016you} achieve remarkable performance when their training is done on the fully labeled large-scale datasets, which include both instance-level annotations (i.e. bounding boxes around each object of interest) and image-level labels (i.e. category of the object enclosed in a bounding box). On the one hand, collecting a dataset with full annotations, especially bounding boxes, can be a tedious and costly process. On the other hand, the object detectors such as R-CNN and YOLO show that their performance is dependent on accessing to such fully labeled datasets. In other words, they suffer from a drop in generalization performance when trained on partially labeled datasets (i.e., containing instances with missing labels)~\cite{wu2018soft,zhang2018weakly,wu2018soft}.
\begin{figure}[t!]
    \centering
    \includegraphics[trim= {0cm 6cm 0cm .2cm}, clip, width=0.5\textwidth]{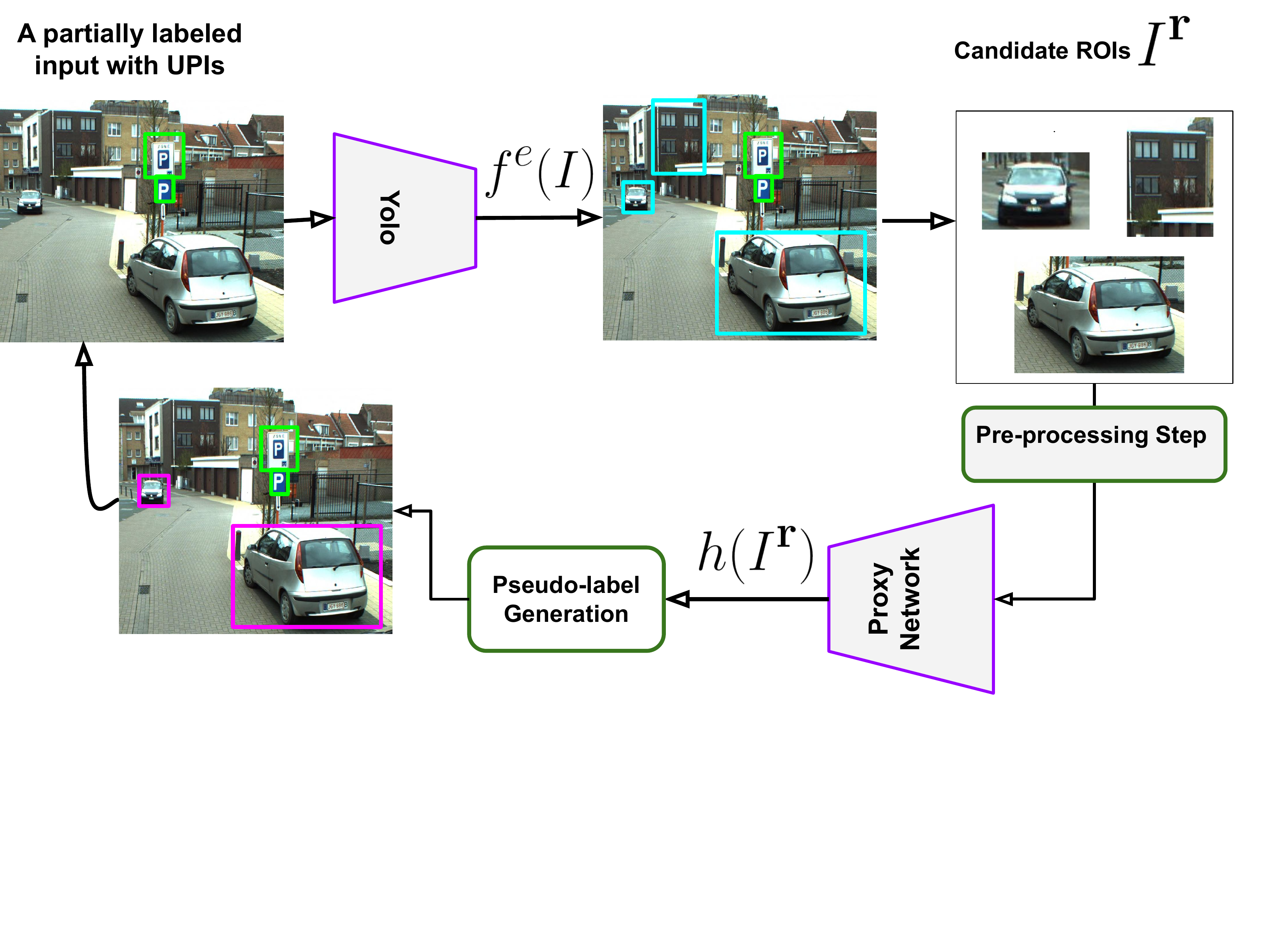}
    
    \caption{Schematic explanation of our proposal for generation of pseudo-labels in a merged dataset. For a given input $I$ with some UPIs (Unlabeled Positive Instance), the bounding boxes (RoIs) estimated by YOLO at training epoch $e$ (i.e. $f^e(I)$) are extracted for a pre-processing step, i.e. to prepare them for the proxy network. Using the proxy network's estimations for the given RoIs, we create pseudo-labels for UPIs allowing YOLO to be trained jointly with the pseudo-labels and the ground truths of the given input.}
    \label{proposed}
\end{figure}
Datasets with missing label instances can occur in several situations, including unintentional errors occurring in the annotation process, partial-labeling policy (we explain it later), and the merged datasets. 
By \emph{merged datasets}, we aim at combining several datasets from similar (or the same) contexts but with disjoint (or partially disjoint) sets of Objects-of-Interest (OoIs), e.g.\ \cite{rame2018omnia}, in order to construct a larger dataset including a wider range of objects, of possibly more variations in their capture and nature (e.g.\ objects of different poses, illuminations, styles, and physical properties). For instance, Kitti~\cite{geiger2013vision} and German Traffic Signs~\cite{Houben-IJCNN-2013} are datasets with two disjoint sets of OoIs that could be merged to cover a wider spectrum of the objects appearing on roads. 

Such merged datasets can facilitate the training of an integrated object detector, which in turn can potentially lead to a significant reduction of time and computational cost. training and inferring from a unified object detector on a merged dataset is more effective in terms of memory and computational resources, compared to training several object detectors, each for one of the constituting datasets. This is specially appealing for the embedded devices with limited computational resources (e.g.\ self-driving cars) as they need to make the inference decisions in real-time manner. In addition, training a unified model circumvents the need to combine decisions made by the various models, which can be tricky and lead to sub-optimal solutions. Finally, merging datasets and training a unified object detector on it can pave the path toward the development of an universal object detector (e.g.~\cite{wang2019towards}).
Despite the great potential of merging-dataset for the reduction of the computational cost and annotation burden, it unfortunately results in missing-label instances as some OoIs in one dataset might are not labeled in other datasets.

Many modern object detectors that are trained on a partially labeled dataset, e.g. a merged dataset, induce inferior generalization performance than those trained with the fully labeled ones~\cite{wu2018soft,xumissing}. Regardless the type of object detectors, the small number of labeled instances in a partially-labeled dataset is one reason for such performance degradation. The anther reason is rooted from false negative training signals arising from the Unlabeled Positive Instances (UPIs). Inspired by~\cite{wu2018soft}, we later elaborate in Sec.~\ref{impact} how these UPIs can mislead training of an object detector, particularly YOLO.

To augment the training size of such partially labeled datasets, Weakly Supervised Learning (WSL) methods~\cite{zhang2018weakly,xumissing,bilen2016weakly,diba2017weakly} have been proposed to generate pseudo-labels for some UPIs by leveraging the \emph{image-level labels}, which are only available in the datasets that are annotated by "partial annotation policy". To reduce the annotation cost, this policy aims to annotate only one instance of each object if it is presented in a given image and the rest ROIs with the same object category are left unlabeled. Although this policy creates a dataset with some missing instance-level labels (i.e. \emph{bounding-box annotations}), it assures that all the images have their true image-level labels (i.e. object category).
Unfortunately, such WSL methods can not be simply employed for the merged datasets in order to mitigate the missing-label instances since in Such datasets, both image-level and instance-level annotations are missed.

To mitigate the performance degradation in faster R-CNN trained on partially labeled datasets (e.g. OpenImagev3~\cite{openimagev3} as it is labeled by "partial annotation policy"), Wu et al.~\cite{wu2018soft} propose to ignore the false training signals arising from UPIs (i.e. false negative). To this end, they discard the gradients created by the RoIs that have small or no overlap with any ground truths. Although this simple approach can remove the false negative training signals by UPIs, correcting them, instead of ignoring them, can further improve generalization performance, particularly for the merged dataset. In other words, to benefit from the differences in the appearance of objects in the merged dataset as well as to obtain a well-generalized unified object detector, it preferably should be trained on all of the positive instances, both the labeled and the unlabeled (UPIs) ones. In~\cite{rame2018omnia}, the authors proposed to generate a set of pseudo-labels for UPIs in the merged dataset by using several different object detectors, where each is trained separately on an individual dataset in the merged one. Finally, another unified object detector is trained on the offline set of pseudo-labels and the ground-truth. However, generating such offline set of pseudo-labels by this approach is computationally expensive (in term of time, memory, and GPU) as both training and label-inference of various distinct object detectors leads to a computational burden.

In this paper, we aim at enhancing the generalization performance of an object detector, when it is trained on a merged dataset, through augmenting it with the on-the-fly (online) generated pseudo-labels for some UPIs. For that purpose, we propose a computationally inexpensive and general \emph{training framework} for training a single detector (e.g. YOLO) while simultaneously creating pseudo-labels for some UPIs. Fig.~\ref{proposed} illustrates the pipeline of our proposed method. We deploy a pre-trained proxy CNN for flagging which YOLO's predicted bounding-boxes contain UPIs, then generate the "object" and "class" pseudo-label for them (Alg~\ref{main_Alg}). In other words, if the proxy network classifies them as one of the pre-defined object classes (OoIs), their pseudo-labels are created to being included in the training phase of the object detector, otherwise, they are discarded from contributing in the training.
Inspired by~\cite{abbasi2019OOD,hendrycks2018deep}, we use a CNN with an explicit rejection option as the proxy network, in order to either classify a given RoI into one of the pre-defined classes or reject it as a not-of-interest object. 

\section{Background}
\begin{figure}[h!]
    \centering
    \includegraphics[width=0.5\textwidth, trim=0cm 5cm 0cm 1cm,clip=true]{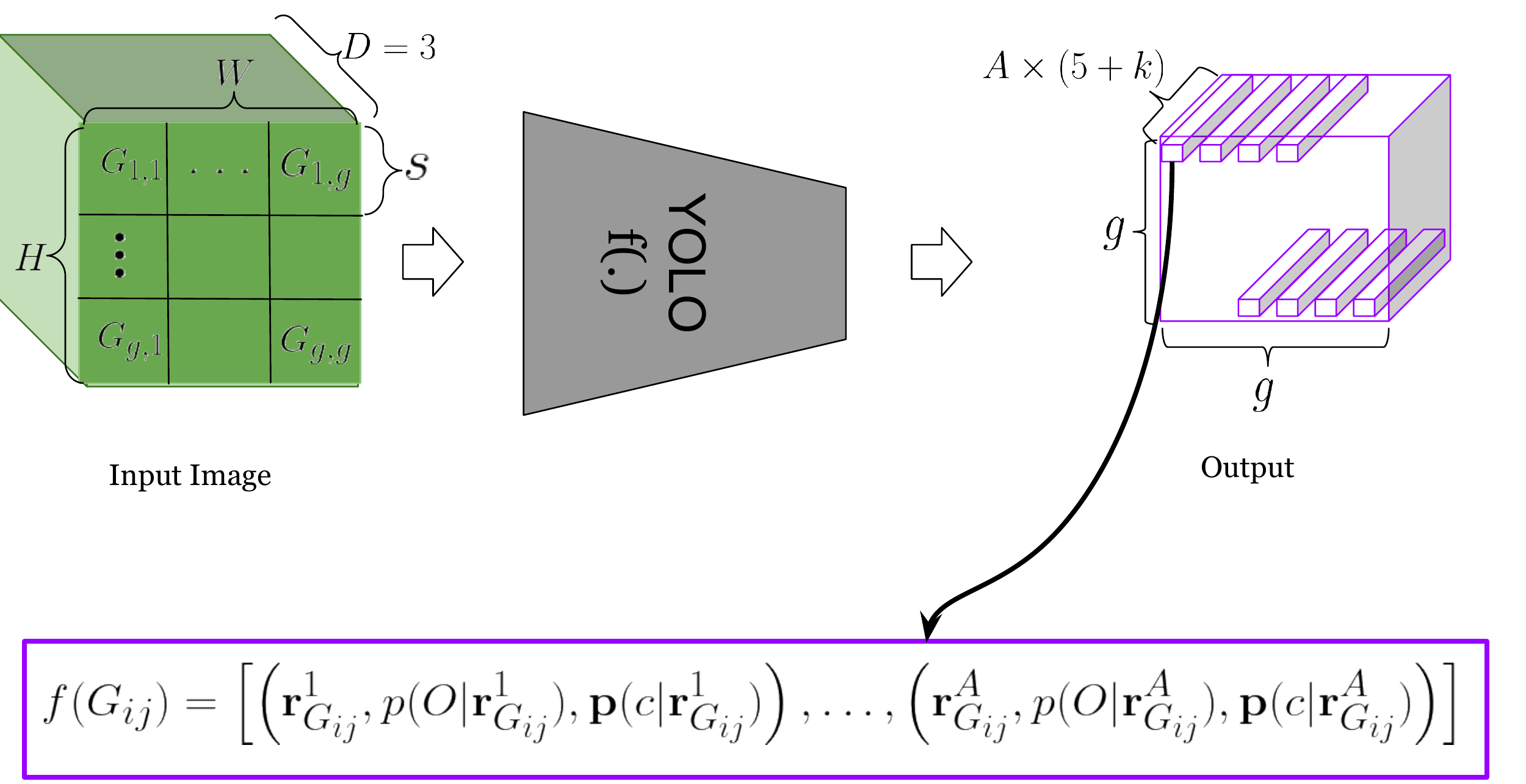}
    \caption{The output of YOLO is a tensor of size $\left[g,g,A,5+K\right]$. }
    \label{yolo}
\end{figure}

YOLO divides a given image $I$ into $g\times g$ grids, then for each grid $G_{ij}$, it estimates $A$ different bounding-boxes, where each of them is a $5+K$-dimensional vector, encompassing the estimated coordinate information of the box (i.e. $\mathbf{r}^a_{G_{ij}}=\left[\hat{x}^a_{G_{ij}},\hat{y}^a_{G_{ij}}, \hat{w}^a_{G_{ij}}, \hat{h}^a_{G_{ij}} \right]$), the objectiveness probability (i.e. $p(O|\mathbf{r}^a_{G_{ij}})$), and a $K$-dimensional vector as the probabilities over $K$ object categorizes (i.e. $\textbf{p}(c|\mathbf{r}^a_{G_{ij}})\in[0,1]^K$) with $a\in\{1,\dots A\}$. Therefore, the output of YOLO will be a tensor of size $\left[g,g,A,5+K\right]$ (Fig.~\ref{yolo}). Moreover, for each grid ${G_{ij}}$, a set of pre-defined bounding-boxes (called anchors) with different aspect ratios and scales is considered. YOLO learns to estimates the bounding-boxes w.r.t these pre-defined anchors.

\section{The Impact of Missing-label Instances on Performance }\label{impact}

As stated earlier, the missing-label instances (called UPIs) can cause false negative signals in the training. In the following, we demonstrate how the missing-label instances can negatively impact on the performance (i.e. mean average of precision ) of object detectors, particularly YOLO and faster R-CNN.

YOLO computes the "object loss" for all anchors of all the grids, whether they contains any ground-truths or not. In other words, if an anchor has no ground-truth (a large IoU overlap, e.g. IoU(anchor, ground-truth) $>\theta$), its true "object" label is zero, and is one, otherwise. This can cause a false negative signals by UPIs. More specifically, during training of YOLO, the detector may be able to localize correctly a UPI, thus the objectiveness probability of its corresponding anchor is $p(O)\sim1$, but since it has no associated ground-truth label, it is given a true "object" label zero, i.e. $t=0$. This forces the network to learn it as a negative or not-interesting object even thought the network can correctly localize and recognize such a UPI (Unlabeled Positive Instance). Ultimately such a false negative signal from a UPI can confuse the network since the LPIs of the same object category forces the network to learn it as an object of interest while the UPIs from the same object category encourage the network to learn it as an not-interesting (negative) object. Therefore, such false negative signals can cause a drop in the performance of YOLO. \emph{Note that the "class" and "coordinate" losses are ignored for the anchors that have small IoU overlap with a ground truth, i.e. IoU$\leq\theta$.} Thus, while UPIs can not contribute in the training through their "class" and "coordinate" losses. The loss functions of YOLO are defined in Appendix~\ref{yolo_losses}.

Similarly, in the faster R-CNN, a UPI can penalize the network incorrectly if its anchor has an IoU (with a ground truth) smaller than a given threshold, e.g. $\theta_1=0.3$. More precisely, the true objectiveness label of an anchor involving a UPI (i.e. $t$) will be set to zero when its IoU overlap is small ($<\theta_1$). Then, although the RPN maybe can localize correctly the UPI as a positive instance (i.e. $P(O)\sim1$), the "object" loss is incorrectly penalizing the PRN by forcing it to learn the UPI as a a negative instance. Thus, such false negative signals from UPIs can intervene with the true positive signals from LPI, leading to a drop in performance of the faster R-CNN. However, interestingly, according to the faster R-CNN described in~\cite{ren2015faster}, if an anchor involving a UPI has no high IoU overlap (e.g. $>theta
_2=0.7$) nor small IoU (e.g. $< \theta_1=0.3$), then such an anchor and its probable corresponding UPI, will be ignored to contribute in the training. Comparing to YOLO, this simple condition in the faster R-CNN may reduce the probability of the false negative signals by UPIs (those that have $\theta_1\leq$IoU$\leq\theta_2$).

Consequently, to mitigate the performance degradation of these object detectors, we require to reduce the false negative signals that are created by their "object" loss. To alleviate this challenge, one possible way is to discarded these false negative signals likewise~\cite{wu2018soft}. However, one can improve further the performance if these false negative signals can be corrected through generating pseudo-labels for them. This can not only reduce the number of false negative signals but also can increase the number of labeled instance, which they together can finally enhance the performance of the object detector.

\section{Proposed Method}

\begin{algorithm*}[t]
\begin{algorithmic}[1]
    \Require{$f^{e}(\cdot)$ object detector at training epoch $e$; $h(\cdot)$ pre-trained proxy network; $I$ given input image with its associated ground-truth bounding-boxes $\mathcal{R}^*$ (i.e. their coordinate information) ; $\theta_{1},\theta_{2}$ and, $\beta$ as hyper-parameters.}

    \Ensure{$S^{e}$, pseudo-labels of $I$ at time $e$}
    \NoDo
    \NoThen
    
    \State{ $S^{e}=\emptyset$}
    \vspace{.2cm}
    \State{ $\left[[\mathbf{r}^{e}_{1}, ~p(O|\mathbf{r}^{e}_{1}), ~\mathbf{p}(c|\mathbf{r}^{e}_{1})], \dots, [\mathbf{r}^{e}_{Ag^2}, ~p(O|\mathbf{r}^{e}_{Ag^2}), ~\mathbf{p}(c|\mathbf{r}^{e}_{Ag^2})] \right]= f^{e}(I)$}
    
    \vspace{.2cm}
    
    \State $\mathcal{R}^e = \{\mathbf{r}_1^e, \dots, \mathbf{r}_{Ag^{2}}^{e}\}$
    \vspace{.2cm}
    \State $\mathcal{P}^e=\{ ~\mathbf{p}(c|\mathbf{r}^{e}_{1}), \dots  ~\mathbf{p}(c|\mathbf{r}^{e}_{Ag^2})\}$
    \vspace{.2cm}

     \State{$\mathcal{B}=\{\emptyset\}$}
    \vspace{.2cm}
    \For{$\mathbf{r} \in \mathcal{R}^{e}$}
    
     \If {\text{IoU}(~$\mathbf{r}$,~$\mathcal{R}^*$) $\leq \theta_{1}$}\Comment{\parbox[t]{.5\linewidth}{To skip generation of pseudo-labels for the estimated ROIs with a large IoU overlap with a ground-truth from $\mathcal{R^*} $. } }
    
    \State{ $\mathcal{B} \longleftarrow \mathcal{B} \cup \{\mathbf{r}\}$}
    \EndIf
    \EndFor

    \State $\mathcal{B}\longleftarrow$ pre-processing step ($\mathcal{B}$)
     \For{$\mathbf{r} \in \mathcal{B}$}
     \State{ $\{I^{\mathbf{r}_1},\dots, I^{\mathbf{r}_m}\}$ = patch-drop($I^{\mathbf{r}}$)} \Comment{\parbox[t]{.5\linewidth}{Create $m$ copies of RoI extracted by $\mathbf{r}$, i.e. $I^{\mathbf{r}}$.}}
    \vspace{0.2cm}
    \State $\bar{h}(I^{\mathbf{r}}) = \frac{1}{m+1}\left( h(I^{\mathbf{r}})+\sum_{i=1}^{m} h(I^{\mathbf{r}_{i}})\right)$
    \vspace{.2cm}

    \If {$\arg\max \bar{h}(I^{\mathbf{r}})\neq K+1$ \& $\max_{\{1,\dots,K \}} \bar{h}(I^{\mathbf{r}}) \geq \theta_{2}$}
    \vspace{.2cm}
    
    \State $\mathbf{\Tilde{p}}(c|\mathbf{r})=\beta \cdot \mathbf{p}(c|\mathbf{r})+(1-\beta) \cdot \bar{{h}}(I^{\mathbf{r}})$\Comment{\parbox[t]{.5\linewidth}{ $\mathbf{p}(c|\mathbf{r})\in\mathcal{P}^e$is the class estimation for the given $\mathbf{r}$. }}
    
\vspace{.2cm}

    \State $ \Tilde{p}(O|\mathbf{r}) = \max_{\{1,\dots,K \}} \bar{h}(I^{\mathbf{r}})$

    \vspace{.2cm}
    
    \State $S^{e}\longleftarrow S^{e} \cup [\hat{r},~\mathbf{\Tilde{p}}(c|\mathbf{r}),~\Tilde{p}(O|\mathbf{r}) ] $
    \EndIf
    \EndFor
    \caption{Pseudo-label Generation Algorithm}
    
    \label{main_Alg}
\end{algorithmic}
\end{algorithm*}

We introduce our framework to handel missing-label instances when the underlying object detector is YOLO, however it can also be adapted for the faster R-CNN.
During training of YOLO, it is likely that some existing UPIs are localized correctly. However due to the lack of a ground truth label for them, they may adversely contribute in the training of YOLO, inducing a drop in performance. We propose to generate psuedo-label for them. 
The estimated RoIs by YOLO at training epoch $e$ are evaluated to check whether they actually contain a positive unlabeled object or not. To achieve this, our framework incorporates a pre-trained proxy network~\cite{abbasi2019OOD}, denoted by $h(\cdot)$, into the training process of YOLO. Indeed, the proxy network maps the current estimated RoIs $\mathbf{r}\in\mathcal{R}^e$ of a given image $I$ (denoted by $I^{\mathbf{r}}$ ), into a $K+1$-dim vector of probabilities over $K+1$ classes, i.e. $h(I^{\mathbf{r}})\in[0,1]^{K+1}$, where $\{1,\dots,K\}$ denotes the class of $K$ positive objects (OoI) and $K+1$-th (extra) class is for any uninterested (negative) objects. Note that to enable $h$ for processing these RoIs with different aspect ratios, we exploit a Spatial Pyramid Pooling (SSP) layer~\cite{he2015spatial} after the proxy's last convolution layer.

To achieve this proxy trained, we can leverage from the readily accessible datasets that contain the samples from not-interested-objects (we call them Out-of-Distribution --OoD--
samples) along with the labeled samples containing OoIs (a.k.a.\ in-distribution samples). Recently, some promising results of OoD training have been reported for developing  a robust object recognition classifiers~\cite{abbasi2019OOD,hendrycks2018deep,Meinke2020ood}, semantic segmentation models~\cite{bevandic2018discriminative}, as well as for overcoming catastrophic forgetting~\cite{lee2019overcoming}.

Using the coordinate information $\mathbf{r}$ provided by YOLO, an estimated RoI $\mathbf{r}$ is extracted from an image $I$.    
To avoid re-labeling the RoIs containing a ground truth, only those that have a small or no overlap with any of the ground truth annotations (line 4--6 of the algorithm~\ref{main_Alg}) are processed. Before feeding these extracted RoIs to the proxy network, they should be pre-processed by the following procedure. 

\subsection{Pre-processing Step}
To allow $h(\cdot)$ processes the RoIs in a mini-batch style, we perform this pre-processing step. Indeed, 
training of $h$ with the mini-batch SGD on the input samples with different aspect ratio sizes is challenging since Python libraries such as Pytorch do not allow the input samples with various sizes to be stacked in one batch. To address this issue, we can think of padding the inputs with the largest aspect ratio size in the batch, but this can destroy the information of the smallest inputs (since these images can be dominated by an extremely large pad of zeros). To tackle this, in each training epoch of $h$, we load the samples with similar (close) aspect ratios in one batch and pad them with zeros, if needed, to achieve a batch of samples with the equal aspect ratio size. To implement this, all of the training samples are clustered by their widths and heights using $k$-means method. Then, the centers of these clusters serve as the pre-defined aspect ratios to load the batches accordingly. Therefore, \emph{in the pre-processing step, at the test time of $h$}, all the input instances to $h$ (i.e. RoIs) should be padded with zeros, if needed, in order to keep their size equal to their nearest centers (line 7 of Algorithm~\ref{main_Alg}).

\subsection{Pseudo-label Generation}
 Inspired by~\cite{singh2017hide}, we make use of patch-drop at the test time of $h$ in order to estimate the true class of a given RoI more accurately. In the patch-drop, the given RoI is divided into $s\times s$ patches, then randomly drop one of them to create a new version of the RoI. In our experiments, we apply patch-drop with $s=3$ for $m=2$ times on a given RoI to create $m$ versions of $I{\mathbf{r}}$, i.e. $\{I^{\mathbf{r}_1},\dots I^{\mathbf{r}_m} \}$ (line 11 in Alg~\ref{main_Alg}). We then feed them as well as the original RoI $I{\hat{r}_j}$ to the proxy network for estimating the probability over $K+1$ classes as follows:
 \begin{equation}
 \label{barh}
     \bar{h}(I^{\mathbf{r}}) = \frac{1}{m+1}\left(h(I^{\mathbf{r}})+\sum_{i=1}^{m} h(I^{\mathbf{r}_{i}})\right).
 \end{equation}
 This trick leads to more calibrated confidence prediction, especially for some hard-to-classify RoIs, as the proxy network $h$ predicts each version of $I^{\mathbf{r}_i}$ differently (to different classes). This indeed allow us to reduce the number of false positive instances, thus the creation of more accurate pseudo-labels. Indeed, using a threshold on the predictive confidence $\bar{h}(.)$ (i.e. $\theta_2$ in the algorithm), the RoIs with low confidence prediction are dropped to continue the pseudo-label generation procedure.  If the proxy network \textbf{confidently} classifies the given RoI into one of $K$ classes, its pseudo class probability $\Tilde{\mathbf{p}}(cls|{\mathbf{r}})$ is computed as follows:
 \begin{equation}
    \mathbf{\Tilde{p}}(c|\mathbf{r})=\beta \cdot \mathbf{p}(c|\mathbf{r})+(1-\beta) \cdot \bar{{h}}(I^{\mathbf{r}}),
     \label{pseudo-class}
 \end{equation}
 where $\mathbf{p}(c|\mathbf{r}), \bar{{h}}(I^{\mathbf{r}})\in [0,1]^K$ are respectively the estimated class probabilities by YOLO at training epoch $e$ and the proxy network $h$ for the given RoI $I^{\mathbf{r}}$. 
 Finally, we set the probability of object for the given RoI $\mathbf{r}$ as
 $\Tilde{p}(O|\mathbf{r})=\max_{k=1}^{K} \bar{h}(I^{\mathbf{r}}).$

\begin{figure*}[h!]
    \centering
    \subfigure{\includegraphics[width=0.25\textwidth,trim=0cm 0cm 0cm 0cm, clip=true ]{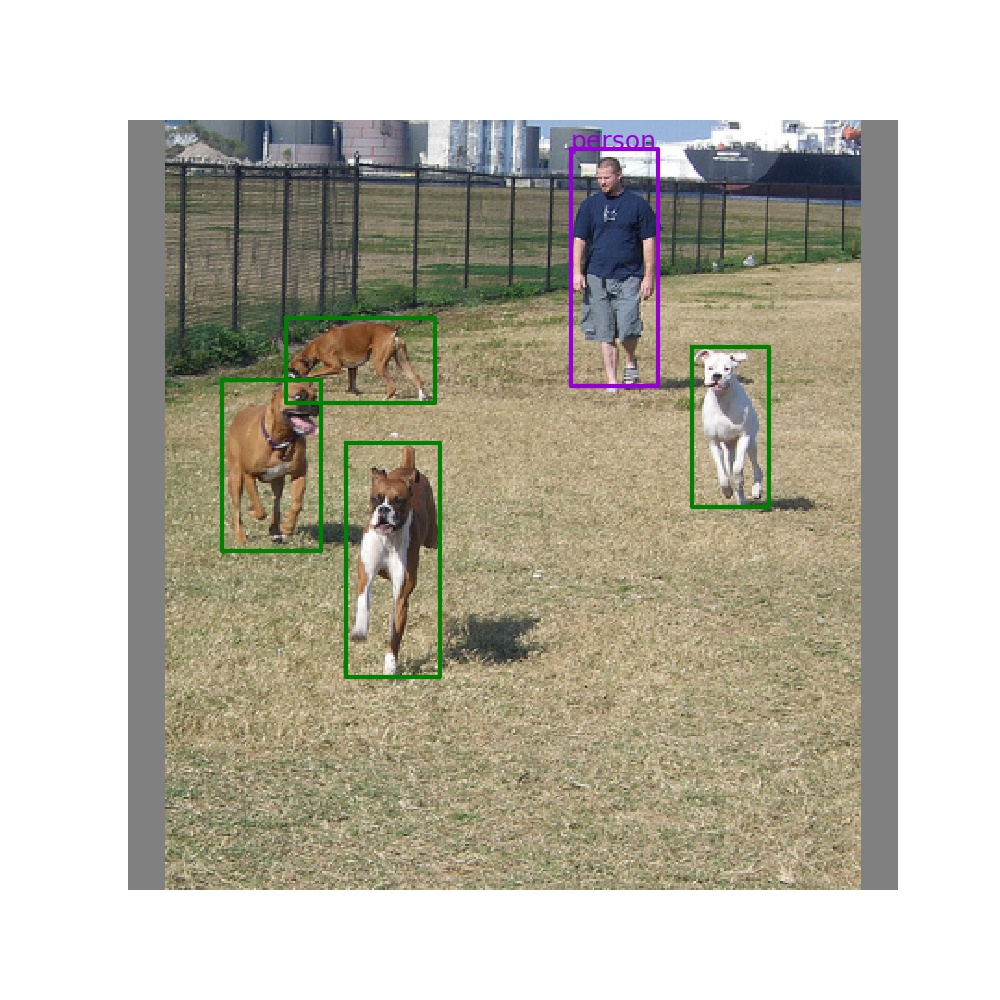}}~
    \subfigure{\includegraphics[width=0.25\textwidth,trim=0cm 0cm 0cm 0cm, clip=true ]{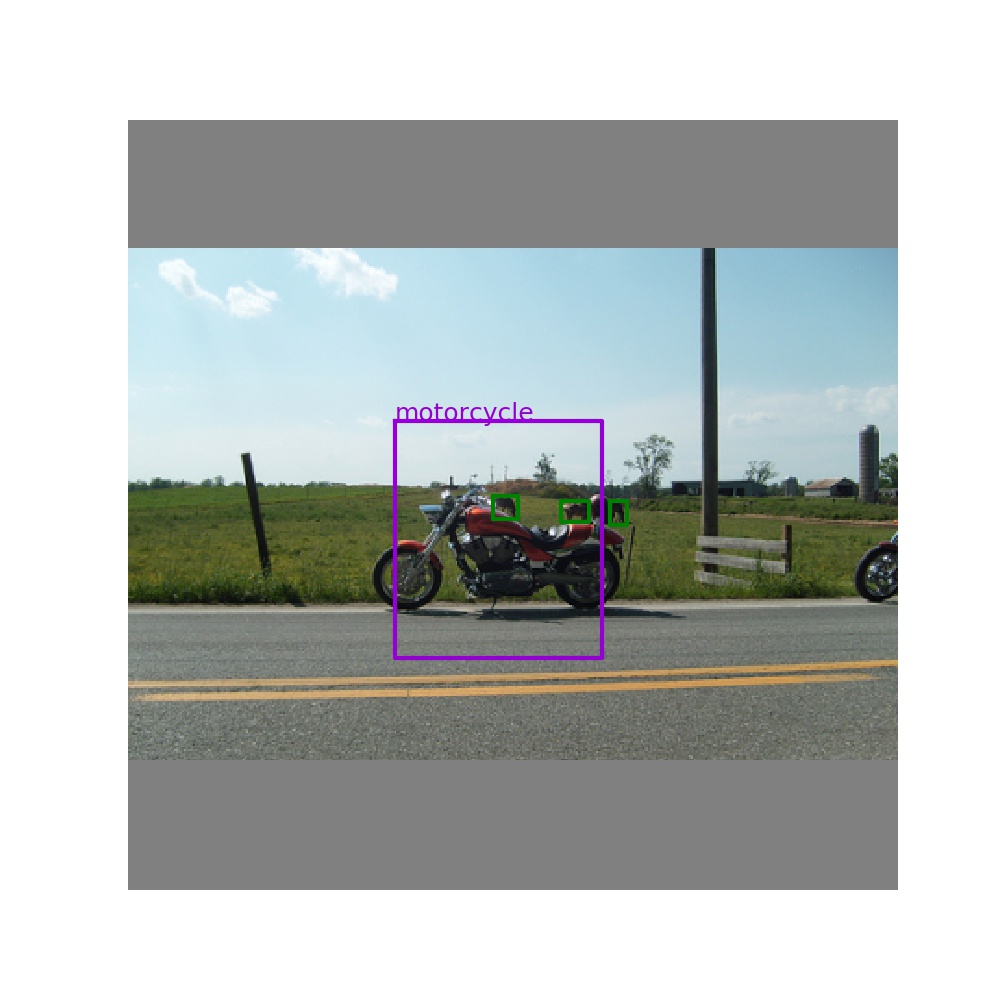}}~
    \subfigure{\includegraphics[width=0.25\textwidth,trim=0cm 0cm 0cm 0cm, clip=true ]{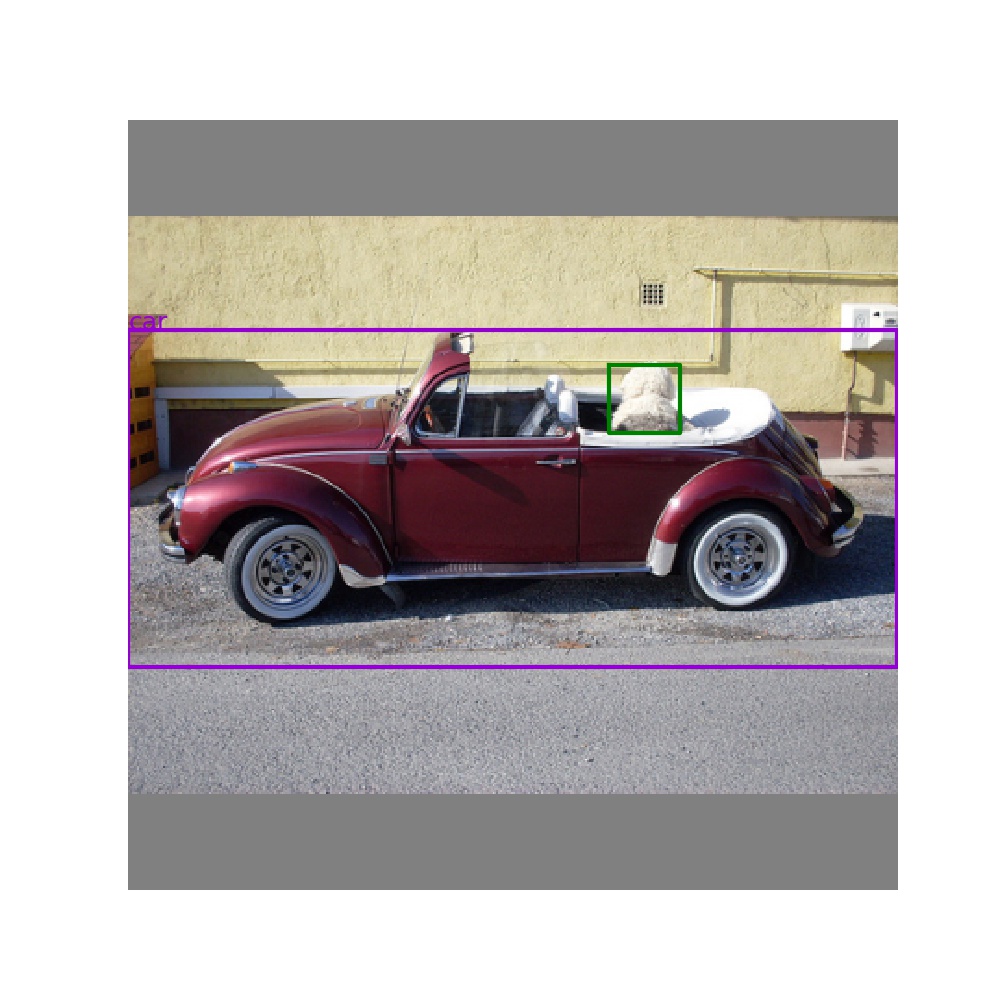}}~\subfigure{\includegraphics[width=0.25\textwidth]{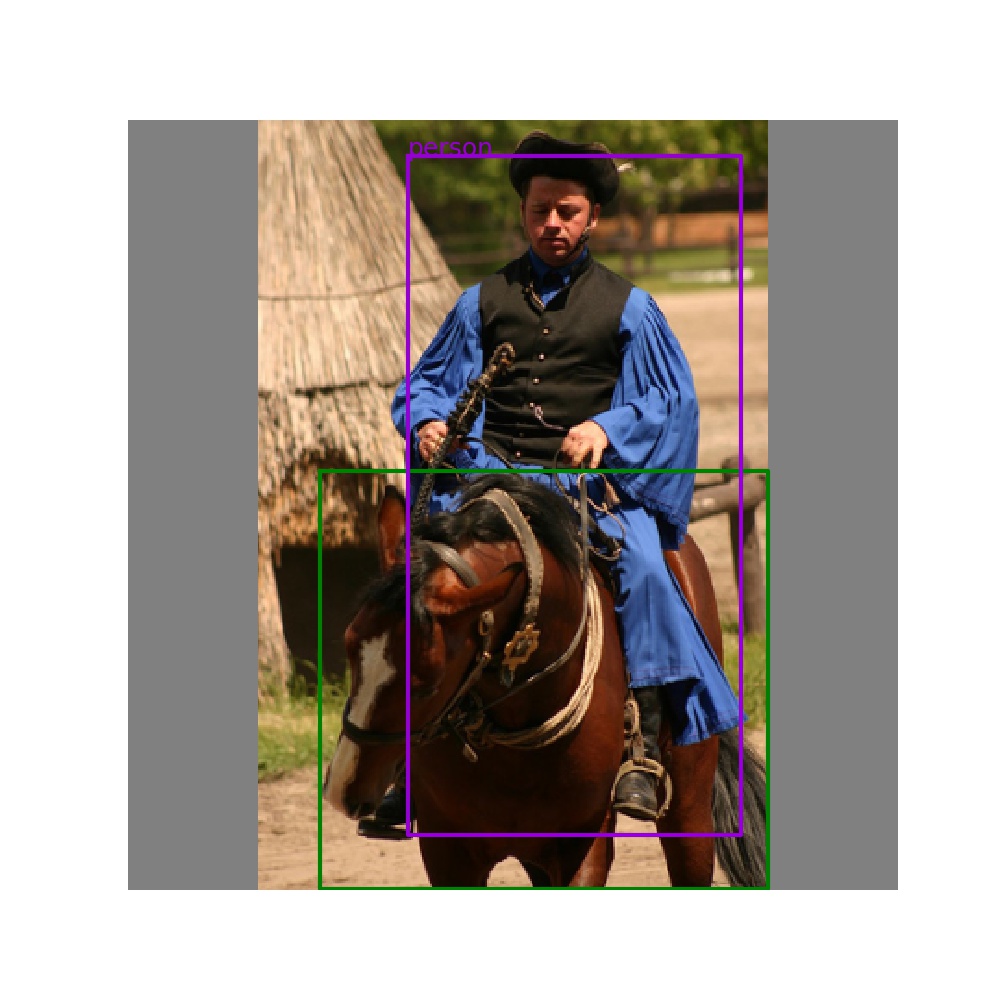}}
    
    \caption{Violet bounding boxes are our pseudo-labels generated during training of Yolo while the green bounding boxes are the ground-truth labels in dataset $D'_{S}$ (i.e. the merged dataset from VOC2007 and VOC2012 with disjoint sets of classes.)}
    \label{vis_exp1}
\end{figure*}

 To compute the loss for the pseudo-class label, we use KL-divergent between the "class" pseudo-label $\mathbf{\Tilde{p}}(c|\mathbf{r})$ and its estimation $\mathbf{p}(c|\mathbf{r})$ by YOLO. Similarly, the "object" loss for the pseudo "object" label $\Tilde{p}(O|\mathbf{r})$ is computed by a binary cross-entropy. Finally, these two new losses for the pseudo-labels, i.e. $KL\left(\mathbf{\Tilde{p}}(c|\mathbf{r})|| \mathbf{p}(c|\mathbf{r})\right)$ and $\text{BCE}(\Tilde{p}(O|\mathbf{r}), p(O|\mathbf{r}))$,  are added to the conventional loss functions of YOLO, which are defined in Appendix~\ref{yolo_losses}.

 \section{Experiments}

To simulate a merged dataset, we create two datasets with two disjoint sets of classes from VOC2007 with $S_A$=\{cat, cow, dog, horse, train, sheep\} and VOC2012 with $S_B$=\{car, motorcycle, bicycle, aeroplane, bus, person\}. One dataset, called $D_{S_{A}}$, gathers the samples from VOC2007 that are containing one of the objects of interest in $S_A$ (dropping the annotations from other set of classes $S_B$, if there are any in $D_{S_{A}}$). Similarly, another dataset  $D_{S_{B}}$ is made of the images from VOC2012 containing one of objects in $S_B$. Then, these two datasets are merged to produce a merged dataset $D'_{S} = D_{S_{A}} \cup D_{S_{B}}$ with total classes of $S =S_A\cup S_B$. In addition, a fully labeled dataset $D_S$ from the union of VOC2007 and VOC2012 are formed, where all the instances belonging to $S$ are fully annotated. The missing label rate of $D'_{S}$ (the merged dataset) with respect to $D_S$ is $48\%$.

\begin{table}[h!]
\centering

\begin{tabular}{cccc}
\hline
Object & \multicolumn{3}{c}{mAP@0.5}\\
 &Baseline& Ours& Upper-bound \\
\hline
\hline
Cat & 74.79 & \textbf{77.2} & \color{gray}{82.04} \\
Cow & 48.27 & \textbf{55.6} & \color{gray}{69.70} \\
Dog & 52.71 &\textbf{62.0} & \color{gray}{78.70} \\
Horse & 18.68 & \textbf{23.7}&\color{gray}{82.51} \\
Train & \textbf{58.36} & 57.7 & \color{gray}{79.18} \\
Sheep & 57.77 & \textbf{65.1} & \color{gray}{72.45} \\
Car & 77.67 & \textbf{78.3} & \color{gray}{83.87} \\
Motorbike & 68.23 & \textbf{72.4} & \color{gray}{79.82} \\
Bicycle & 69.98 & \textbf{72.1}  & \color{gray}{79.00} \\
Aeroplane & 59.96 & \textbf{62.6} & \color{gray}{71.29} \\
Bus & 65.26 & \textbf{71.2} & \color{gray}{78.83} \\
Person & 71.32 & \textbf{72.0} & \color{gray}{78.30} \\
\hline  
Avg & 60.25 & \textbf{64.2} & \color{gray}{77.97} \\
\end{tabular}
\caption{Performance (i.e. mAP) of different Yolos on the test set of VOC2007 with fully labeled instances from classes $\mathcal{S}=\mathcal{S}_A \cup \mathcal{S}_B$. Baseline is the trained Yolo on the merged dataset (voc2007+voc2012) with missing-label instances ($D'_{S}$), ours is Yolo trained on the augmented dataset $D'_{S}$ with our generated pseudo-labels, and the upper-bound is the Yolo trained on voc2007+voc2012 with \textbf{fully} annotated instances ($D_{S}$).}
\label{VOC-exp1}
\end{table}

As the proxy network, we adopt Resnet20~\cite{he2016deep} by placing a SPP (Spatial Pyramid Pooling) layer after its last convolution layer to enable it to process the inputs with various aspect-ratio sizes. To train this network, we utilize MSCOCO~\cite{lin2014microsoft} training set by extracting all the ground truth bounding boxes belonging to one of the classes in $S=S_A \cup S_B$, and all other ground truth bounding boxes \emph{not} belonging to $S$ are used as OOD samples (labeled as class $K+1$). The hyper-parameters of our algorithm are set to $\beta=0$ (in Eq.~\ref{pseudo-class}), $\theta_1= 0.5$ (to remove RoIs having a large overlap with ground truth, line 4--6 of Algorithm), and $\theta_2=0.8$ (the threshold on the prediction confidence of the proxy network for the given RoIs).

In Fig.~\ref{vis_exp1}, we demonstrate the pseudo labels generated by our proposed method for some UPIs in $D'_{S}$. 
In Table~\ref{VOC-exp1}, we compare mAP@0.5 of three Yolos, where they are respectively trained on $D'_S$ (baseline), on augmented $D'_S$ by our pseudo-labels (Ours), and finally on fully labeled dataset $D_S$. As it can be seen, training a YOLO on $D'_{S}$ (with a $48\%$ rate of missing labels) leads to a $\approx\!17\%$ drop in mAP@0.5, compared to the same YOLO when it trained on the fully-labeled dataset ($D_S$). Ours enhances mAP of YOLO trained on the merged dataset $D_S'$ by $~4\%$ (on average) by augmenting $D'_S$ by pseudo-labels for some of UPIs, thus their false negative signals are eliminated.

\section{Conclusion}
With the goal of training an integrated object detector with the ability of detecting a wide range of OoIs, one can merge several datasets from similar context but with different sets of OoIs. While merging multiple datasets to train an integrated object detector has some promising potentials, from reducing the computational and labeling costs to enjoying from a wider spectrum of variations (suitable for domain-shift), many missing label instances (Unlabeled Positive Instances) in the merged dataset cause a performance degradation. To address this issue, we propose a general training framework for simultaneously training an object detector (e.g. YOLO) on the merged dataset while generating some on-the-fly pseudo-labels for UPIs. Using a pre-trained proxy neural network, we generate a pseudo label for each estimated RoI if the proxy network confidently classifies it as one of its pre-defined interested classes. Otherwise, we exclude it from contributing in training of the object detector. By a simulated merged dataset using VOC2007 and VOC2012, we empirically show that YOLO trained by our framework achieves a higher generalization performance, compared to the YOLO trained on the original merged dataset (with the missing-labels). This achievement is the result of augmenting the merged dataset with our generated pseudo-labels for UPIs.



%
\bibliographystyle{IEEEtran}
\bibliography{ref}

\appendix
\section*{{Loss function of YOLO}}
\label{yolo_losses}

Each training sample is $({I}_i, \{\mathbf{t}_{i}^{*1}, \dots \mathbf{t}_{i}^{*j}\})$, where ${I}_i$ is an input image and $\mathbf{t}_{i}^{*j}$ is the $j$-th ground truth bounding-box associated with $i$-th image ($i\in\{1,\dots N\}$). Each ground truth bounding-box is $\mathbf{t}_{i}^{*j}=[\mathbf{r}^{*j}_{i}, {k}_{i}^{*j}]$ with $\mathbf{r}^{*j}_{i}=[{x}_{i}^{*j}, {y}_{i}^{*j}, w_{i}^{*j}, h_{i}^{*j}]$ and ${k}^{*j}_{i}\in\{1,\dots,K\}$ is the object category. The coordinate information of the \emph{center} of $j$-th ground-truth and its corresponding height and width w.r.t the image are  ${x}_{i}^{*j}, {y}_{i}^{*j}, w_{i}^{*j}, h_{i}^{*j}\in[0,1]$, respectively. From now on, we drop the indices from the ground-truth and their estimations for the simplicity reasons.

Contrary to the coordinate information of the ground-truth, i.e. $\mathbf{r}^*=\left[x^*, y^*, w^*, h^*\right]$, that of estimated bounding-box by YOLO, i.e. $\mathbf{r} = \left[\hat{x},\hat{y}, \hat{w}, \hat{h}\right]$, are relative to their corresponding gird (grid-orientation). To have the ground-truths and the estimations in the same coordinate-system, the predicted bounding-box is transferred to image's coordinate system as follows:

\begin{align}
    b_{\hat{x}} & = x_{G_{ij}} +\hat{x}\\
    b_{\hat{y}} &= y_{G_{ij}}  +\hat{y}\\
    b_{\hat{w}}^a & = w^a_{G_{ij}} \exp^{(\hat{w})} \\
    b_{\hat{h}}^a &= h^a_{G_{ij}} \exp^{(\hat{h})},
    \label{coord-transfer}
\end{align}

where $x_{G_{ij}}, y_{G_{ij}}$ are the coordinate of the top-left corner of grid ${G_{ij}}$ w.r.t the image, and $w^a_{G_{ij}}, h^a_{G_{ij}}$ are the width and height of $a$-th anchor of the given grid.

  For each grid $G_{ij}$ with $i,j\in\{1, \dots, g\}$ (Fig.~\ref{yolo}), we compute the "class" and "coordinate" losses only if the IoU between a ground-truth, e.g. $\mathbf{r^*}$, and at least one of the grid's anchors $\mathbf{A}^a_{G_{ij}}$ is larger than a pre-defined threshold $\tau$, otherwise its "class" and "coordinate" losses are zero (ignored). \textbf{Note if a grid has several anchors that have large IoU ($>\tau$) with a ground-truth, then the anchor with the largest IoU is solely contribute for computing these losses.}

 For a give $G_{ij}$, let $a' = \argmax_{a}\left( \text{IoU}( \mathbf{r^{*}},\mathbf{A}^{a}_{G_{ij}})\right)$, the "class" loss, i.e. multi-class cross-entropy, computes the difference between the estimated class probabilities, i.e. $\mathbf{p}(c|\mathbf{r}^{a'}_{G_{ij}})$ and the true class $k^*$, which encoded by $\mathbf{p^*}(c)$; that is a one-hot $K$-dimensional vector with its $k^*$-th element equals to one ($\mathbf{p^*}(c=k^*)=1$)~\footnote{Instead cross-entropy for the class predictions, the authors~\cite{redmon2018yolov3} used binary cross-entropy loss for each of $K$ classes.}:

\begin{equation}
\resizebox{.5\textwidth}{!}{
        $\mathcal{L}_{cls} (\mathbf{p^*}(c),\textbf{p}(c|\mathbf{r}^{a'}_{G_{ij}})) =\left\{
        \begin{array}{cc}
             \log p(c=k^*|\mathbf{r}^{a'}_{G_{ij}})& \text{if} ~~\max_{a}\left( \text{IoU}( \mathbf{r^*},\mathbf{A}^a_{G_{ij}})\right)>\tau   \\
             0 & Otherwise.
        \end{array}\right.$}
  \end{equation}

    \begin{equation}
    \resizebox{.5\textwidth}{!}{
    \begin{math}
    \mathcal{L}_{coor} \left(\left[x^*, y^*, w^*, h^*\right], \left[b_{\hat{x}}, b_{\hat{y}}, b_{\hat{w}}^{a'}, b_{\hat{h}}^{a'}\right]|G_{ij}\right)=\left\{
    \begin{array}{cc}
        \begin{aligned}[t] &(x^*-b_{\hat{x}})^2+(y^*-b_{\hat{y}})^2+\\&
         (w^*-b_{\hat{w}}^{a'})^2+(h^*-b_{\hat{h}}^{a'})^2
         \end{aligned} 
         & \text{if} ~~\max_{a}\left( \text{IoU}( \mathbf{r^*},\mathbf{A}^a_{G_{ij}})\right)>\tau   \\
         0 & \text{Otherwise}.
    \end{array}\right.
    \end{math} 
    }
    \end{equation}
In addition, for each grid $G_{ij}$, we compute the "object" loss, i.e. binary cross-entropy, measures the loss on the estimated objectiveness probability for its anchor that has the maximum IoU overlap with a ground-truth and its IoU$( \mathbf{r^*},\mathbf{A}^a_{G_{ij}})>\tau$. To give true "object" label to this anchor, the value of $t^a_{G_{ij}}=1$. For the remaining anchors of the grid $t^a_{G_{ij}}=0$. If none of the anchors of $G_{ij}$ have a large IoU overlap (>$\tau$), then the true "object" label all of them is zero, i.e. $ t^a_{G_{ij}}=0~~ \forall a\in \{1,\dots, A\}$.
\begin{equation}
\label{obj-loss}
\resizebox{0.5\textwidth}{!}{

        $\mathcal{L}_{obj}( t^a_{G_{ij}}, p(O|\mathbf{r}^a_{G_{ij}})) =t^a_{G_{ij}}\log p(O|\mathbf{r}^a_{G_{ij}}) + (1-t^a_{G_{ij}})\log(1-p(O|\mathbf{r}^a_{G_{ij}}))$}
\end{equation}

\textbf{It should be emphasized that the "object" loss is computed for all the grids (i.e. all the anchors of all the grids), whether they contain a ground-truth or not, while the "coordinate" and "class" losses for the grids that have no ground-truth are not computed (since these losses are always zero for such grids, by the definition). }
 Finally, all the above loss functions are weighted summed to define the total loss of YOLO. The weights are set so that the contributions of the losses are balanced.
\begin{equation}
\begin{aligned}
    \mathcal{\mathbf{L}} = &\lambda_{cls}\sum_{G_{ij}}\mathcal{L}_{cls} \left(\mathbf{p^*}(c),\textbf{p}(c|\mathbf{r}^{a'}_{G_{ij}})\right)+\\
    & \lambda_{coor}\sum_{G_{ij}}\mathcal{L}_{coor} \left(\left[x, y, w, h\right],\left[b_{\hat{x}}, b_{\hat{y}}, b_{\hat{w}}^{a'}, b_{\hat{h}}^{a'}\right]|G_{ij}\right)+\\
    &\lambda_{obj}\sum_{G_{ij}}\sum_{a=1}^{A}\mathcal{L}_{obj} \left(t^a_{G_{ij}}, p(O|\mathbf{r}^a_{G_{ij}}))\right)
\end{aligned}
\label{total_loss_yolo}
\end{equation}

\end{document}